\documentclass{article}

\usepackage{times}
\usepackage{graphicx}
\usepackage{subfigure}

\usepackage{natbib}

\usepackage{algorithm}
\usepackage{algorithmic}

\usepackage{hyperref}
\usepackage[accepted]{icml2016}

\icmltitlerunning{A Taxonomy and Library for Visualizing Learned Features in CNNs}

\begin{document}

\twocolumn[
\icmltitle{A Taxonomy and Library for Visualizing Learned Features\\
           in Convolutional Neural Networks}

\icmlauthor{Felix Gr\"un}{felix.gruen@tum.de}
\icmladdress{Technische Universit\"at M\"unchen, Germany}
\icmlauthor{Christian Rupprecht}{christian.rupprecht@in.tum.de}
\icmladdress{Technische Universit\"at M\"unchen, Germany, Johns Hopkins University, Baltimore, USA}
\icmlauthor{Nassir Navab}{navab@cs.tum.edu}
\icmladdress{Technische Universit\"at M\"unchen, Germany, Johns Hopkins University, Baltimore, USA}
\icmlauthor{Federico Tombari}{tombari@in.tum.de}
\icmladdress{Technische Universit\"at M\"unchen, Germany, University of Bologna, Italy}

\icmlkeywords{feature visualization, CNN, convolutional neural networks, machine learning, ICML}

\vskip 0.3in
]

\begin{abstract}
Over the last decade, Convolutional Neural Networks (CNN) saw a tremendous surge in performance. However, understanding what a network has learned still proves to be a challenging task. To remedy this unsatisfactory situation, a number of groups have recently proposed different methods to visualize the learned models. In this work we suggest a general taxonomy to classify and compare these methods, subdividing the literature into three main categories and providing researchers with a terminology to base their works on. Furthermore, we introduce the FeatureVis library for MatConvNet: an extendable, easy to use open source library for visualizing CNNs. It contains implementations from each of the three main classes of visualization methods and serves as a useful tool for an enhanced understanding of the features learned by intermediate layers, as well as for the analysis of why a network might fail for certain examples.
\end{abstract}

\section{Introduction}
\label{sec:Introduction}

In recent years, research in the field of artificial neural networks (ANN) has made tremendous progress in many diverse areas \citep{Taigman:2014,Hannun:2014,He:2015}. Fueled by GPU accelerated learning  and the development of regularization techniques like dropout \citep{Hinton:2012,Srivastava:2014}, activation functions like rectified linear units (ReLU) \citep{Glorot:2011}, LeakyReLUs \citep{Maas:2013}, and parametric ReLUs \citep{He:2015}, and larger labeled datasets \citep{Deng:2009,Lin:2014}, natural image classification with convolutional neural networks (CNN) has become a \textit{flagship example} \citep{Yosinski:2015} for the advancements in supervised learning using neural networks.

Yet, understanding the learned models is still an unsolved problem. Even the researchers who developed new techniques to improve CNN performance relied largely on trial and error \citep{Zeiler:2013}. Historically, the inner units of a network were regarded as a black box \citep{Yosinski:2015}, appropriately affording them the name \textit{hidden layers}. It is therefore not surprising that feature visualization for neural networks is a very young area of research.

The first approaches aimed at visualizing CNNs were made in 2013, first by \citet{Zeiler:2013} and only one month later by \citet{Simonyan:2013}. Since then, a number of visualization methods have been proposed. Gaining an overview of the differences and similarities between the methods and the advantages and disadvantages of each has become a time consuming exercise. However, our careful analysis of the different methods for feature visualization shows that they can be grouped into just three classes. Together, these three classes comprise a novel taxonomy for feature visualization methods, which will provide a terminology in which to discuss existing research and from which to develop new directions and techniques.

Furthermore, we introduce the FeatureVis library\footnote{The library is freely available at: \url{http://campar.in.tum.de/Chair/FeatVis}} for the MatConvNet toolbox \citep{Vedaldi:2015}, which contains implementations for several state-of-the-art visualization methods to provide researchers and practitioners with an easy way to use the different methods for visualization in their own projects. Upon acceptance, the library will be publicly accessible and open source. It works with all CNNs built from the layers available with the MatConvNet toolbox and can easily be extended to incorporate custom layer types. The library can aid in the improvement of existing network architectures and the search for better models, as well as in understanding their performance.

\section{A Taxonomy for Feature Visualization Methods}
\label{sec:Taxonomy}
Our proposal of a taxonomy for feature visualization methods consists of three classes which we refer to as \textit{Input Modification} methods, \textit{Deconvolutional} methods, and \textit{Input Reconstruction} methods. These classes are characterized by both, the goals for which the methods have been developed and the algorithms which they use.

\subsection{Input Modification Methods}
\textit{Input Modification} methods are visualization techniques which modify, as the name suggests, the input and measure the resulting changes in the output or intermediate layers of the network. These methods see the network (or all the layers before the layer of interest) as a black-box and they are designed to visualize properties of the function this black-box represents. The resulting visualizations can be used to analyze the importance and locality of features in the input space. A prime example for this class of visualization methods is the \textit{Occlusion} method by \citet{Zeiler:2013}.

To find the areas of the input image which are most important to the classification result, \citet{Zeiler:2013} systematically cover up portions of the input image with a gray square and measure the change in activations. If an area of the image which is vital to the classification is covered up, the activations change noticeably. Covering up parts of the image which are unimportant to the computed activations, on the other hand, results only in a small difference. If the image is occluded systematically by moving the gray square with a fixed step width from left to right and top to bottom across the whole image, the resulting changes in activation form a heat map showing the importance of different areas of the input image.

\citet{Zhou:2014} extend the Occlusion method proposed by \citet{Zeiler:2013} by using a patch of randomized pixel values for the occlusion instead of a mono colored gray square. Since CNNs are very sensitive to edges, a mono colored area might contribute in unpredictable ways to the network output. Randomizing the pixel values reduces the risk of accidentally introducing new features into the image, which could be picked up by the CNN.

\subsection{Deconvolutional Methods}
In contrast to \textit{Input Modification} methods, \textit{Deconvolutional} methods see the network as a white-box and use the network structure itself for their visualizations. The common denominator amongst the different methods in this class is the idea to determine the contribution of one pixel of the input image by starting with the activation of interest and iteratively computing the contribution of each unit in the next lower layer to this activation. In this way, by moving backwards through the network until the input layer is reached, the contribution values for each pixel can be obtained, which together form a visualization of the features that are most relevant to the activation of interest.

\begin{figure}[t]
\begin{center}
\vskip 0.2in
\centerline{\includegraphics[width=\columnwidth]{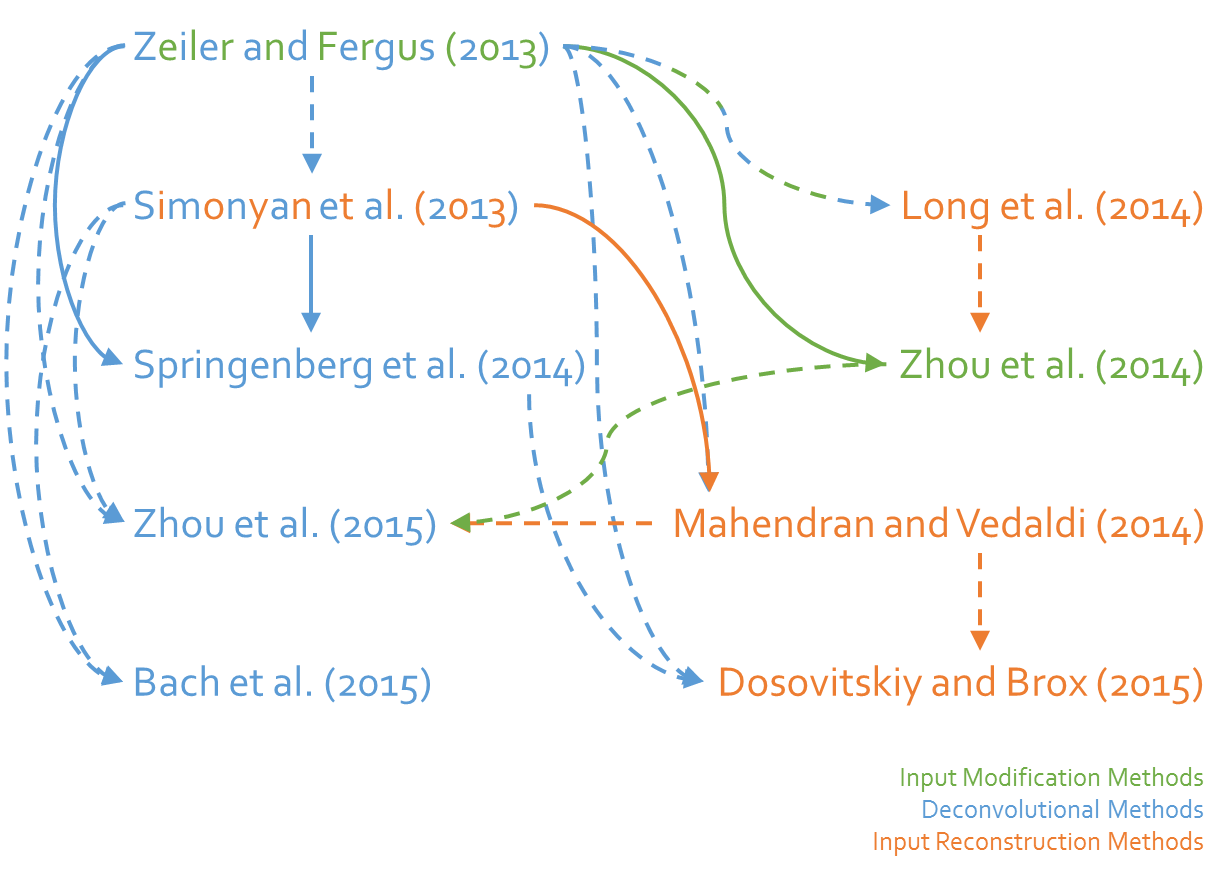}}
\caption{Influences between authors of papers on feature visualization for CNNs. Dashed lines indicate that one author cited another, while solid lines indicate that one author build upon and extended the work of another. The ordering is top down by publication year.}
\label{fig:Influence Between Authors}
\end{center}
\vskip -0.2in
\end{figure}

The first methods of this class were proposed by \citet{Zeiler:2013} and \citet{Simonyan:2013}. \citet{Zeiler:2013} built on the work of \citet{Zeiler:2011} and used a multi-layered Deconvolutional Network (\textit{Deconvnet}) to project the activations from the feature space back to the input space. Deconvnets were originally proposed as a technique for unsupervised learning of CNNs, but can also be used to reverse the path of excitatory stimuli through the network to reveal which pixels and patterns of the input image are responsible for the observed activations.

\citet{Simonyan:2013} build on work by \citet{Baehrens:2010} in the context of Bayesian classification to present a variation of the Deconvnet approach. They reason that the derivative of the class score in the input space defines the relative importance of the input pixels for the resulting class score. Intuitively, this is equal to finding those pixels that have to be changed the least for the greatest effect on the class score of interest. The underlying assumption is that the pixels that make up an object of a class in the input image have a greater contribution than pixels that constitute unrelated objects. To find the derivative of the class score with respect to the input image \citet{Simonyan:2013} backpropagate the class scores from the last layer down through the network up to the input layer.

In the following year \citet{Springenberg:2014} improved upon the previous two methods. They designed a CNN using only convolutional layers. To visualize the features learned by the resulting network they looked to the Deconvnet technique of \citet{Zeiler:2013}. Central to this approach of visualization are discriminative patterns encoded in the max pooling layers of the network. They contain the information on which pixels in the input space are of the highest importance for the resulting feature activations. Since the CNN developed by \citet{Springenberg:2014} does not have max pooling layers, the original approach would not work.

\begin{figure}[t]
  \begin{center}
    \vskip 0.2in
    \centerline{\includegraphics[width=\columnwidth]{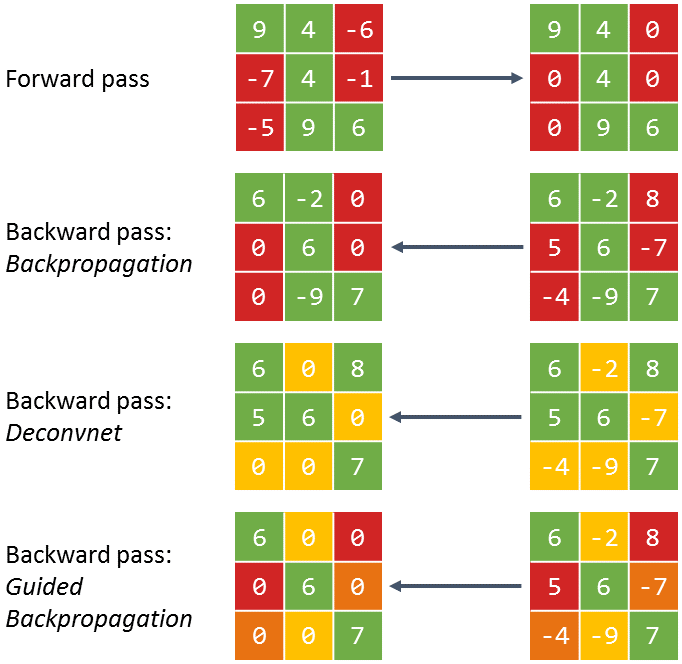}}
    \caption{Different ways in which the pass through a ReLU layer affects contribution values for the Deconvnet method, Backpropagation, and Guided Backpropagation. The forward pass through the ReLU layer is shown for comparison.}
    \label{fig:ReLU pass}
  \end{center}
  \vskip -0.2in
\end{figure}

Building on the work of \citet{Zeiler:2013} and \citet{Simonyan:2013}, \citet{Springenberg:2014} generalized the Deconvnet approach to work with networks that do not have max pooling layers and developed a new technique which enhanced the clarity and sharpness of the projection, even for networks that do. They named the resulting algorithm \textit{Guided Backpropagation}.

  \begin{figure*}[p]
    \begin{center}
      \centerline{\includegraphics[height=0.95\textheight]{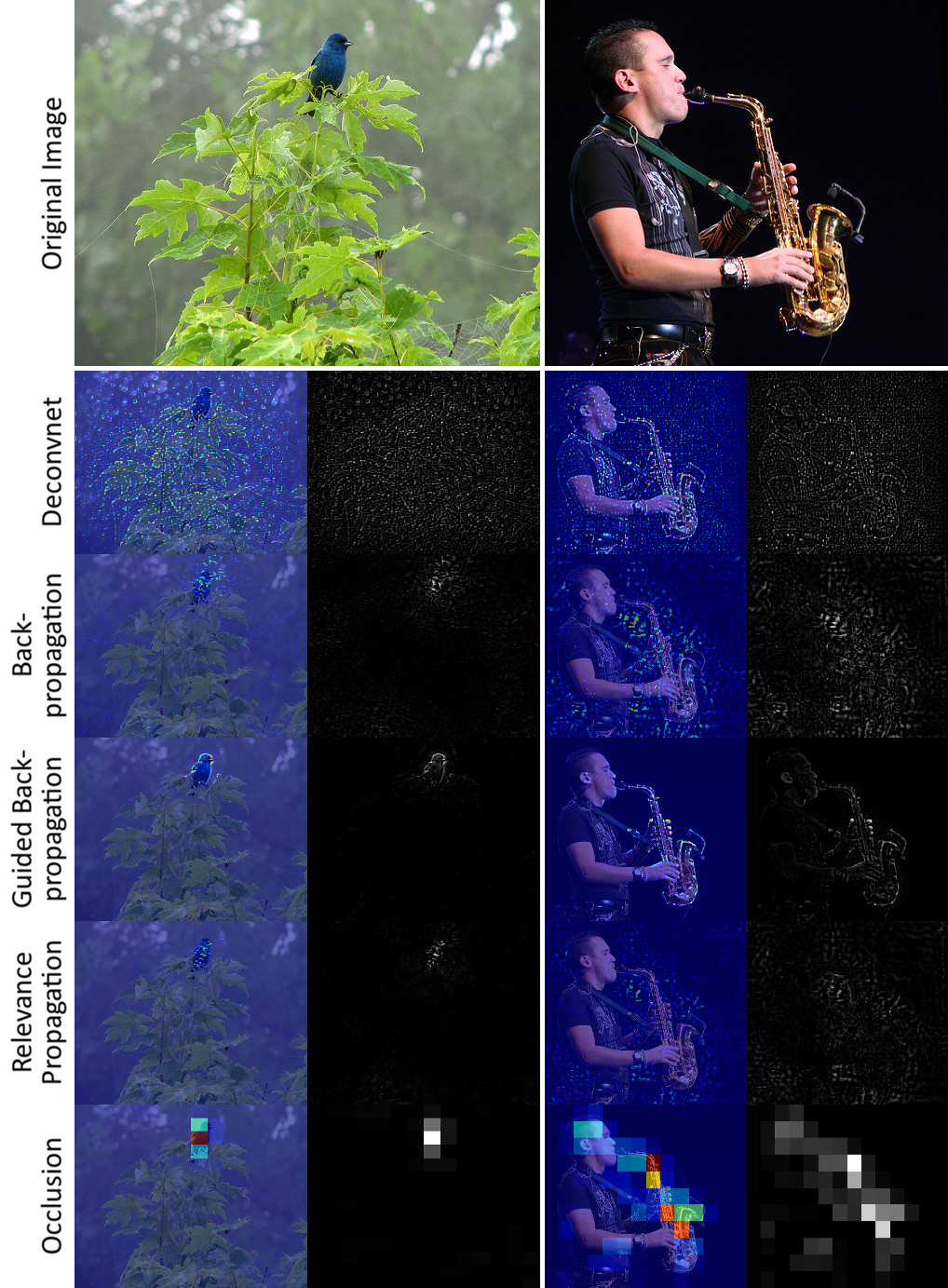}}
      \caption{Comparing different visualization methods for two images. Both images where classified correctly with high confidence by the network. The image classes are: \textit{indigo bunting, indigo finch, indigo bird, Passerina cyanea} and \textit{sax, saxophone}. Visualizations for the respective image class using a VGG-16 network and an epsilon value of 0.001 for the Relevance Propagation method.}
      \label{fig:Visualizations comparison}
    \end{center}
  \end{figure*}

Unlike the approaches presented so far, which specifically deal with the visualization of CNNs, \citet{Bach:2015} aim at finding a general approach for visualizing classification decisions of nonlinear classifiers. To attain this goal they use pixel-wise decomposition to visualize the contributions of each pixel to the final classification decision. In their model, the inputs to the final classification layer have a certain relevance to that layer's activations and their inputs, in turn, have a certain relevance for their activations. This way, relevance flows from the final layer of the network down through the layers to the input image, where it can be mapped to pixels to show the relevance of each pixel for the classification result.

Despite the differences in the initial idea and the language used for description, careful analysis of this approach revealed that the algorithm which \citet{Bach:2015} developed can be compared to the Deconvnet, Backpropagation of activations and Guided Backpropagation approaches described above. In fact, the main differences between the different approaches falling in the \textit{Deconvolutional} methods class are the different ways in which they propagate the contribution values through ReLU and convolutional layers.

While Backpropagation uses the information from the original activations of the lower layer to decide which values to mask out during the pass through a ReLU layer, the Deconvnet approach uses the deconvolved activities from the higher layer. In other words, while Backpropagation uses information from the lower layers and the input image to determine which of the pixel values were important in computing the activations of the higher layers, the Deconvnet approach uses (gradient) information from the higher layers to determine which of the pixels from the input image had a beneficial effect on the activations of the units of interest. Guided Backpropagation combines the two approaches by masking out the value of any unit that had either a negative activation during the forward pass or a negative contribution value during the backward pass. Figure \ref{fig:ReLU pass} provides an overview over how ReLU layers affect contribution values for the different methods.

The last contribution to the class of \textit{Deconvolutional} methods is made by \citet{Zhou:2015}. Their approach requires a CNN which performs global average pooling on the convolutional feature maps just before the final output layer. By using the spatial information of the activation maps before the pooling operation and the values of the output layer weights they construct a \textit{Class Activation Map} showing which areas of the input image were important for the output class of interest. Specifically, a Class Activation Map is the sum of all activation maps before the average pooling operation weighted by the output layer weights connecting the features to the class of interest.

\subsection{Input Reconstruction Methods}
The methods belonging to the third class, i.e. \textit{Input Reconstruction} methods, follow the idea that reconstructing an input, which either maximally activates a unit of interest or leads to the same activations as a natural image prior, reveals which features are important to the associated filters. Here, we take a closer look at three of the proposed methods from this category.

\citet{Long:2014} developed a new way of visualizing the feature detection abilities of a CNN to find out if CNNs learn interclass correspondence. They built on the \textit{HOGgles} approach \citep{Vondrick:2013} of feature visualization for object detectors, but use replacement with the top-\(k\) nearest neighbors in feature space instead of paired dictionary learning. Intuitively, they replace patches of the original input image with patches from other images that lead to the same activations at the layer of interest. This allowed them to evaluate the extent to which CNNs are able to abstract from the input image and recognize higher level features at different layers in the network.

\citet{Simonyan:2013} on the other hand propose a generative class model visualization method which builds upon work done by \citet{Erhan:2009} in the context of visualizing deep belief networks (DBN) and deep unsupervised auto-encoders. They use gradient descent in the input space to iteratively find an image which maximally activates a unit of choice. To find the gradient, they backpropagate the difference between the actual activation and the desired activation to the input layer without changing the network itself. The resulting image contains all the features that the associated filters selects for.

\begin{figure*}[p]
  \begin{center}
    \centerline{\includegraphics[width=\textwidth]{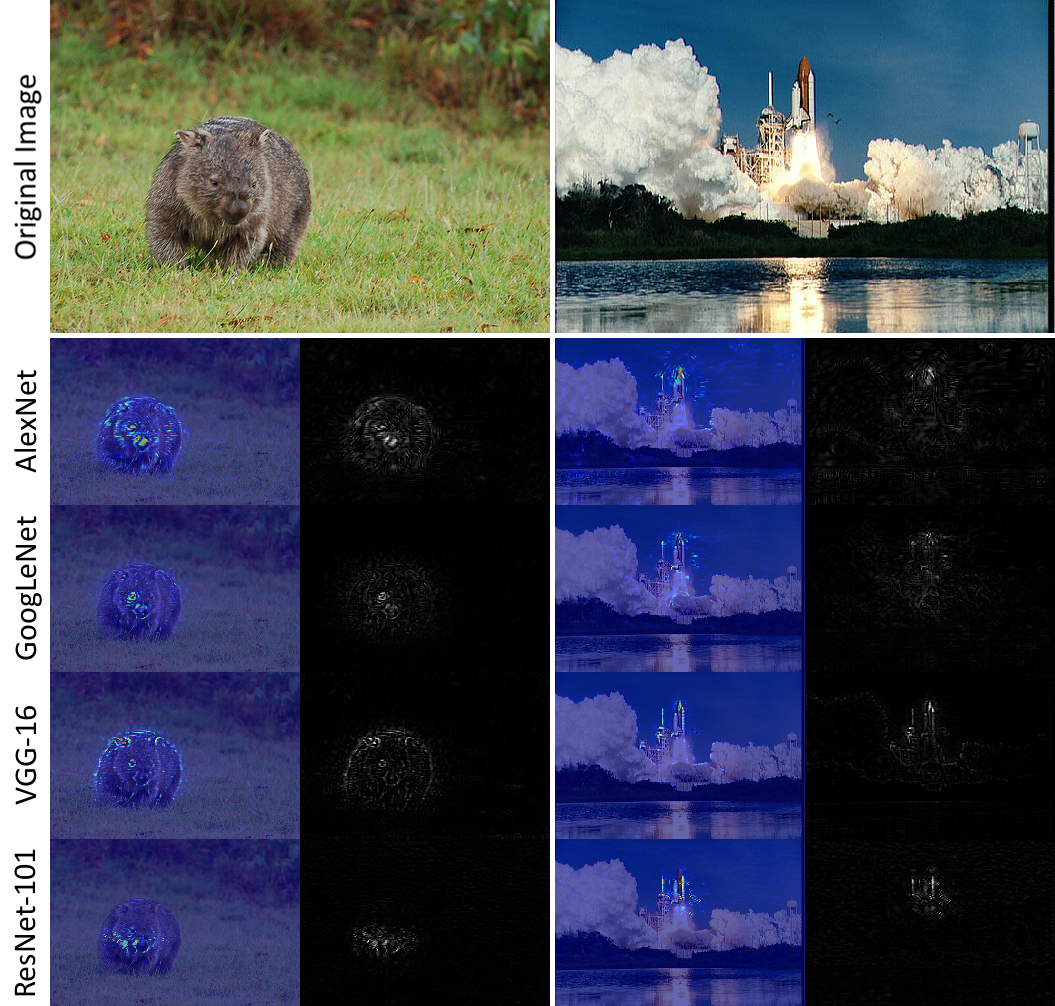}}
    \caption{Comparative visualizations of different networks for two images. Both images where classified correctly with high confidence by all networks. Visualizations for the respective image class using Guided Backpropagation. The image classes are: \textit{wombat} and \textit{space shuttle}.}
    \label{fig:Networks comparison}
  \end{center}
\end{figure*}

In the following year, \citet{Mahendran:2014} tried to invert the image representation encoded in the network output to obtain a reconstruction of the input image. Analyzing which features were present and absent in the reconstructed image would enable a better understanding of the information processed by the network. To this end, they built on the generative class model visualization method of \citet{Simonyan:2013} and made a number of important contributions.

The \textit{Input Reconstruction} method of \citet{Simonyan:2013} and \citet{Mahendran:2014} needs strong regularizers so that the resulting images and the features which they contain can be recognized and interpreted by the human observer. While \citet{Simonyan:2013} used the \(L_2\) norm as a regularizer, \citet{Mahendran:2014} found out, that an \(L_p\) norm with higher values for \(p\) leads to clearer images. In addition to that, they introduce total variation as a second regularizer to be used in conjunction with the \(L_p\) norm. This markedly improved the quality of the end result and lead to more natural looking images.

In contrast to the two methods above using replacement and gradient descent, \citet{Dosovitskiy:2015} use generative networks, i.e. CNNs turned upside down \citep{Dosovitskiy:2014}, to invert the image representation. A generative network receives the representation of an image, e.g. the output of a ``normal'' CNN, as input and computes a reconstruction of the original image. A drawback of this method is that each generative network has to be trained to invert a specific representation before it can be used. This is balanced by the benefit of computing reconstructions several orders of magnitude faster than using gradient descent, after training is complete.

\section{The FeatureVis library}
\label{sec:The FeatureVis library}

To facilitate the use of feature visualization methods for CNNs, we developed an open source library named FeatureVis, built on top of the popular MatConvNet toolbox. MatConvNet implements CNNs for MATLAB and provides different layer types as building blocks that allow for fast prototyping of new CNN architectures \citep{Vedaldi:2015}.

The FeatureVis library is continually growing and already implements functions for many popular feature visualization methods from all three classes. For the deconvolution approach three different methods for the propagation of activations through ReLUs, Deconvnet \citep{Zeiler:2013}, Backpropagation \citep{Simonyan:2013} and Guided Backpropagation \citep{Springenberg:2014}, and two different methods for the propagation of activations through convolutional layers, Backpropagation and Relevance Propagation \citep{Bach:2015}, have been implemented. For the occlusion approach the size of the occluding box and the stride rate can be specified independently for the two dimensions. Experiments with the FeatureVis library showed that the color of the occlusion box matters in determining the importance of occluded features. Therefore, as an improvement over \citet{Zeiler:2013}, the occluded area will either be filled with a color manually chosen by the user or random values as used by \citet{Zhou:2014}. For the input reconstruction approach using gradient descent the user can choose and combine the \(L_p\) norm regularizer \citep{Mahendran:2014,Simonyan:2013} and total variation \citep{Mahendran:2014}.

Figure \ref{fig:Visualizations comparison} shows a comparison of different visualization methods included in the FeatureVis library applied to two example images. The visualizations were created using a VGG-16 network trained on the 1000 classes of the ImageNet Large Scale Visual Recognition Challenge (ILSVRC) \citep{Russakovsky:2015}. The images very clearly show the progress in \textit{Deconvolutional} methods, with Guided Backpropagation offering the sharpest visualizations of the features important for the classification of the input images.

An interesting use-case of the FeatureVis library is the comparison of different network architectures. Figure \ref{fig:Networks comparison} shows visualizations of two images for four different network architectures: AlexNet (ILSVRC top-1 error rate of $41.8$\%), GoogLeNet ($34.2$\%), VGG-16 ($28.5$\%), and ResNet-101 ($23.4$\%).
It is interesting to see that smaller error rates go hand in hand with more sharply focused features contributing to the final classification results. This can be seen best in the black and white visualizations of the pixel contributions for the space shuttle image.

This is a very good example of how visualization methods can facilitate the understanding of the performance of different network architectures. Additionally, it is noteworthy that most visualization techniques are not confined to classification tasks. FeatureVis is agnostic to the loss-layer of the network and can therefore also be used for networks with regression tasks such as human pose estimation, facial landmark detection, semantic segmentation, and depth prediction.

\section{Conclusion}

In this paper we have proposed a taxonomy for feature visualization techniques for deep networks, subdividing the literature into three main classes. For each class, we have described its defining characteristics and summarized the related literature.
Together with the taxonomy we have introduced the open source library FeatureVis for MatConvNet. The library contains implementations of a number of popular feature visualization methods, to be used with any CNN designed using the standard MatConvNet layers with no further setup required. It thus provides a useful tool for the visual analysis of deep models and for direct improvements of a user's network architecture.

While the FeatureVis library can already be used to visualize CNNs using a variety of different methods, more remains to be done. We will focus our efforts on extending the FeatureVis library to incorporate more visualization methods and provide interactive real-time visualizations to assist the comparison of different methods and visualize the influence of different parameters.

\bibliography{sources}
\bibliographystyle{icml2016}

\end{document}